%% file: main.tex
\documentclass[10pt,conference,a4paper]{IEEEtran}
\usepackage[utf8]{luainputenc}
\usepackage{times}
\usepackage{graphicx}
\usepackage{amsmath}
\usepackage{amssymb}
\usepackage{amsthm}
\usepackage{booktabs}
\usepackage{algorithm}
\usepackage{algorithmic}
\usepackage{balance}
\usepackage{enumitem}
\usepackage{subcaption}

\usepackage[hidelinks]{hyperref}

\newcommand{\bgm}{{\sf BackgroundMellow}}
\newtheorem{example}{Example}

\makeatletter

\renewcommand{\thesection}{\arabic{section}}
\renewcommand{\thesubsection}{\thesection.\arabic{subsection}}
\renewcommand{\thesubsubsection}{\thesubsection.\arabic{subsubsection}}
\renewcommand{\thesectiondis}{\thesection.}
\renewcommand{\thesubsectiondis}{\thesubsection.}
\renewcommand{\thesubsubsectiondis}{\thesubsubsection.}

\makeatother

\begin{document}
\title{\bgm: A Multi-Modal Cohesive Framework for Narrative-Driven Rich Cinematic Soundscape Generation}

\author{{\bf Ajitesh Jamulkar and Aritra Hazra}\\Department of Computer Science \& Engineering, Indian Institute of Technology Kharagpur, India.}

\maketitle

\input{sections/1_abstract}
\input{sections/2_introduction}

\input{sections/3_related_work}
\input{sections/4_methodology}
\input{sections/5_results_and_evaluation}
\input{sections/6_future_and_conclusion}

\vfill
\section*{\textbf{Available Resources}}
\smallskip

{\small
\begin{description}[style=nextline,leftmargin=0pt]
 \item[{\bf Code Repository Link:}]
 {\scriptsize \url{https://github.com/anonymous-ismir/BackgroundMellow_ismir.git}}
 \smallskip

 \item[{\bf Experimental Results Link:}]
 {\scriptsize \url{https://docs.google.com/spreadsheets/d/1QyqbvlgpzJ0clwwLy8J5hq-WthJ6IBAN}}
 \smallskip
 
 \item[{\bf Demo Video Link:}]
 {\scriptsize \url{https://www.youtube.com/playlist?list=PLk9zUrtrxbKIFo1hXIi7w1fz7L0m3Qznf}}
\end{description}
}
\clearpage

\balance
\input{sections/7_references}

\end{document}

%% file: sections/1_abstract.tex
\begin{abstract}
 Generating immersive, synchronized and cinematic audio for long-form textual narratives remains a significant challenge in multi-modal AI. While current Text-to-Audio (TTA) frameworks successfully synthesize isolated sound effects, they struggle with narrative cohesion, temporal alignment, and cinematic emotional depth. We present \bgm, a framework that treats story-to-audio generation as a precise orchestration and signal processing problem. This framework is enabled without ground-truth through a master-specialist agent architecture that decomposes text into precise and multi-layered audio cues, generates each category of sounds with suitable specialist model, and superimposes the soundscapes to create a unified and aligned audio segment. Our pipeline is built over Tango2 latent diffusion model for environmental synthesis alongside a novel Cinematic BGM Retriever mined from professional soundtracks. To automate the sound mixing process, we use an NLP based module that predicts precise audio parameters, like start time, duration, and relative loudness, based on the narrative timeline. We further empirically evaluate and show the efficacy of the proposed framework leveraging nearest-neighbor retrieval against a curated dataset of YouTube cinematic trailers to measure temporal synchronization, coverage, and spectral richness.
\end{abstract}

%% file: sections/2_introduction.tex
\section{Introduction}

Recent advancements in Text-to-Audio (TTA) generation, driven by Latent Diffusion Models (LDMs) and flow-matching frameworks, have demonstrated remarkable success in synthesizing high-fidelity acoustic clips. However, a critical frontier remains largely unsolved: the generation of long-form, multi-track narrative soundscapes. Real-world cinematic storytelling is not a monolithic acoustic event; it is a complex, temporally orchestrated mixture of expressive speech, discrete sound effects (SFX), continuous environmental ambience, and contextual background music (BGM).

Current state-of-the-art end-to-end models often excel at isolated sound generation but fundamentally struggle with compositional and acoustic complexity. When prompted to generate simultaneous, overlapping events, they frequently suffer from phase cancellation, acoustic muddiness, and temporal drift. For instance, successfully rendering the narrative prompt, \textit{``It was softly raining as I walked through the forest, where I heard a dog barking,''} requires far more than basic semantic understanding. It demands the cohesive orchestration of a continuous rain ambience, an emotive background score, and a discrete dog bark SFX that triggers at the precise temporal onset of the corresponding spoken word. Maintaining the structural integrity of these diverse acoustic elements and superimposing them with millisecond precision remains a highly complicated and challenging task.

Furthermore, the absence of holistic and rigorous evaluation mechanisms for such multi-stem acoustic narratives presents a fundamental research bottleneck. Cinematic sound design is inherently subjective; a generative model might synthesize a soundscape that utilizes a different musical genre or richer ambient textures than a human baseline, making it equally valid or even superior. Consequently, establishing a rigid, absolute ``ground truth'' dataset for multi-track audio is fundamentally flawed, as it penalizes generative diversity and creative variance. Existing objective metrics, such as Fréchet Audio Distance (FAD)~\cite{fad} or global Contrastive Language-Audio Pretraining (CLAP)~\cite{clap} scores, are heavily optimized for evaluating isolated acoustic events. They are notoriously ill-equipped to assess the temporal synchronization, hierarchical audio ducking, and semantic harmony required for a multi-class cinematic mix.

To bridge this gap, we propose an end-to-end, multimodal cohesive framework, named \bgm, which shifts the paradigm of narrative audio generation from monolithic latent synthesis to parametric orchestration and Digital Signal Processing (DSP). By decoupling the generation of individual acoustic stems from their temporal alignment, we achieve unprecedented control over the multi-track superimposition of cinematic soundscapes. 

The primary contributions of this paper are as follows:
\begin{itemize}
    \item {\em Hierarchical Orchestration Framework.} We develop a Master-Specialist agent architecture that decomposes complex narrative text into precise, multi-layered audio cues, dispatching them to specialized generation and retrieval models (e.g., Tango2~\cite{tango,tango2} , AudioLDM2~\cite{audioldm2}, and a Cinematic BGM Retriever).
    \item {\em Deterministic Temporal Alignment.} We propose a hybrid alignment engine utilizing a fine-tuned Digital Sound Predictor (DSPred) that anchors dynamic SFX, Ambience, and Music directly to word-level timestamps extracted from a Text-to-Speech (TTS) master track, enabling precise acoustic superimposition.
    \item {\em Retrieval-Based Evaluation Metrics.} Addressing the lack of multi-track evaluation standards, we present a novel nearest-neighbor evaluation framework grounded in a curated dataset of professional YouTube cinematic trailers. We introduce custom Semantic Recall and Temporal Intersection-over-Union (IoU) metrics that rigorously evaluate rhythmic pacing and synchronization without penalizing valid generative diversity.
    \item {\em Empirical Evaluation and Benchmarking.} We demonstrate the efficacy of our proposed framework through extensive ablation studies.
\end{itemize}
The rest of the paper is organized as follows. Section~\ref{sec:related_work} reviews the existing works in text-to-audio generation and evaluation. Section~\ref{sec:methodology} details our proposed framework, including the Master-Specialist architecture and the acoustic generation layer. Section~\ref{sec:superposition} further illustrates the superposition and alignment of soundscapes. Section~\ref{sec:eval_metric} introduces the relevant metrics and methods used for evaluations. Section~\ref{sec:result} presents the empirical results and ablation studies. Section~\ref{sec:conclusion} concludes the paper and discusses the future directions.

%% file: sections/3_related_work.tex
\section{Related Work}
\label{sec:related_work}
Our framework addresses the critical limitations of state-of-the-art generative audio models, specifically focusing on the transition from isolated sound generation to complex acoustic orchestration.

\subsection{Latent Diffusion and Long-Form Audio Generation -- The ``Mixture'' Problem}
Latent Diffusion Models (LDMs)~\cite{ddpm}, such as \textbf{AudioLDM}~\cite{audioldm2} and \textbf{Tango2}~\cite{tango,tango2}, alongside foundational music generation systems~\cite{musicgen, musiclm, bach2bach}, represent the standard for high-fidelity text-to-audio generation. While highly proficient at synthesizing isolated sound effects, they inherently struggle with complex, multi-layered mixtures. When prompted to generate simultaneous overlapping events (e.g., speech, rain, and music), monolithic models often produce noisy audio where elements lose spatial separation and fidelity. Consequently, our framework utilizes these models solely as ``specialists'' to generate isolated stems, rather than relying on them to render a complete mix.

Generating coherent long-form audio introduces significant temporal challenges. Concurrent frameworks exploring dynamic soundtracks and multi-agent orchestration~\cite{audiobook_agent, m2m_gen, metabgm}, such as \textbf{AudioStory}~\cite{audiostory}, tackle this by employing LLMs to condition a diffusion transformer (DiT), concatenating generated segments sequentially. While temporally coherent, this end-to-end approach produces a flattened, ``baked'' audio file where dialogue, music, and effects are inextricably fused. 

In contrast, \bgm~frames narrative audio as a parametric orchestration and Digital Signal Processing (DSP) problem. Rather than forcing a generative model to hallucinate a complex mixture simultaneously, our system predicts explicit mixing parameters (start time, duration, and decibel weight) for isolated stems. This superimposition approach preserves high acoustic fidelity---addressing complex alignment challenges noted in recent multi-modal evaluation literature~\cite{fad, clap, align_evaluation}---and enables professional-grade audio ducking, capabilities that end-to-end latent models cannot easily achieve.

\subsection{Expressive Speech Synthesis and Temporal Anchoring}
In cinematic storytelling, the narrator's pacing dictates the temporal structure of the scene. We use expressive Text-to-Speech (TTS) models, such as \textbf{Parler-TTS}~\cite{parler-tts}, to generate the dramatic readings required for narrative audio. Crucially, this TTS output serves as a deterministic master clock for our framework. Unlike purely LLM-driven planners that mathematically estimate timestamps, we extract absolute word-level timestamps directly from the generated speech using robust Automatic Speech Recognition (ASR)~\cite{whisper}. This ensures that superimposed audio cues are grounded to an exact timeline, triggering precisely when the corresponding word is spoken to maximize multimodal alignment.

%% file: sections/4_methodology.tex
\section{Overall Methodology and Architecture}
\label{sec:methodology}
\begin{figure}[h]
\centering
\includegraphics[width=\linewidth]{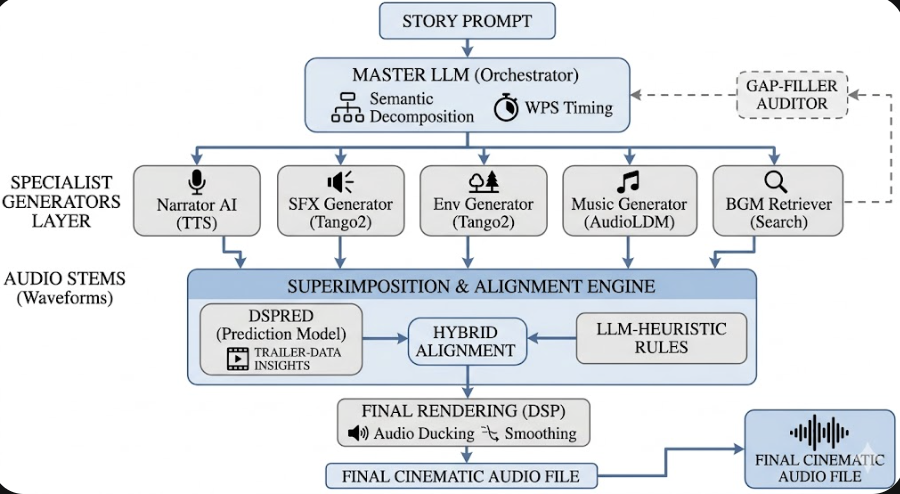}
\caption{Hierarchical Master-Specialist Architecture of \bgm~Framework}
\label{fig:architecture}
\end{figure}

The \bgm~framework operates on a hierarchical Master-Specialist paradigm, effectively decoupling the semantic planning of a narrative from the raw synthesis of its audio components. The architecture is divided into two primary modules: the {\em Specialist Generators} and the {\em Superimposition Engine}, as illustrated in Fig.~\ref{fig:architecture}.

\subsection{The Master Agent: Orchestration and Timing}
Upon receiving a story prompt, the Master Large Language Model (LLM) acts as the orchestrator and performs a semantic analysis of the narrative to decompose the text into a sequence of discrete audio events. Each element in this manifest adheres to an \textit{AudioCue} schema, encompassing:

\begin{itemize}
\item {\tt \textbf{audio\_class}}: The semantic textual prompt defining the sound (e.g., ``heavy rain'', ``distant dog bark'').
\item {\tt \textbf{audio\_type}}: The categorical designation of the sound (SFX, Ambience, Music, or Narrator).
\item {\tt \textbf{start\_time\_ms \& duration\_ms}}: Starting time and duration indicating temporal boundaries for the cue.
\item {\tt \textbf{weight\_db}}: Relative volume adjustment in decibels (dB).
\item {\tt \textbf{fade\_ms}}: Smoothing duration to prevent zero-crossing distortion (defaulted to 500ms).
\end{itemize}

Crucially, the Master Agent mathematically aligns these cues to the narrator's timeline using word-level indexing. The exact initial onset time of any given cue is calculated as:
$$T_{start}=\left(\frac{W_{index}}{WPS}\right)\times1000$$
where $W_{index}$ represents the zero-based positional index of the narrative trigger word and $WPS$ represents the reading speed in Words Per Second. Furthermore, the Master Agent dynamically calculates the \texttt{weight\_db} based on a localized concurrency rule: the relative loudness of a cue is determined by both its semantic intensity and the density of overlapping tracks at $T_{start}$, ensuring a balanced acoustic hierarchy. 

\begin{example} {\bf [Semantic Decomposition]} \label{ex:semantic}
{\em
To illustrate the above orchestration process, let us consider the following horror-story prompt: \textit{``She crept down the pitch-black hallway, gasping as a shadow lunged from the closet.''} The Master Agent decomposes this scene into a dense, multi-track manifest:
\begin{enumerate}
    \item \textbf{Narrator:} \textit{audio\_class:} ``Hushed, breathy whisper...''; \textit{weight:} $0.0$~dB; spanning the full timeline.
    \item \textbf{Ambience:} \textit{audio\_class:} ``Deep, oppressive silence with eerie hum''; \textit{weight:} $-22.0$~dB; anchored at $0$~ms.
    \item \textbf{SFX 1:} \textit{audio\_class:} ``Rhythmic wood floorboard creaks''; \textit{weight:} $-6.0$~dB; triggered early in the scene.
    \item \textbf{Music:} \textit{audio\_class:} ``Tense cinematic rising strings''; \textit{weight:} $-15.0$~dB; initially mapped near the climax.
    \item \textbf{SFX 2:} \textit{audio\_class:} ``Heavy cinematic jump scare impact''; \textit{weight:} $+3.0$~dB; mathematically anchored to the trigger word ``lunged.'' \qed
\end{enumerate}
}
\end{example}

\subsection{Specialist Audio Generation}
Once the cue manifest is generated, the tasks are dispatched in batches to a suite of specialist generation models, significantly optimizing inference latency. Based on audio classes, our framework routes tasks to the following optimal configurations:
\begin{enumerate}
\item \textbf{SFX Generator:} Synthesizes discrete, high-impact sound effects (e.g., footsteps, impacts).
\item \textbf{Environment Generator:} Generates continuous background textures and ambient soundscapes (e.g., eerie hums, rainfall).
\item \textbf{Music Generator:} Composes synthetic musical scores tailored to the scene's emotional context.
\item \textbf{Cinematic BGM Retriever:} A retrieval-augmented module that searches a database of professional movie soundtracks, selecting the track with the highest cosine similarity to the story's emotional tone.
\item \textbf{Narrator AI (Parler-TTS):} A highly expressive Text-to-Speech model that renders the story prompt into voice, guided by specific prosody and environmental descriptors.
\end{enumerate}

\subsection{Overcoming Latent Diffusion Temporal Limits}
A fundamental limitation of latent diffusion models like Tango2 is their fixed-length output generation (typically capped at 10 seconds). To achieve arbitrary durations specified by the \texttt{duration\_ms} parameter, we implement a post-generation boundary conditioning algorithm:
\begin{itemize}
\item \textbf{Truncation via Semantic Alignment (Duration $<$ 10s):} We employ a pre-trained Contrastive Language-Audio Pretraining (CLAP) model to segment the 10-second generated audio into overlapping windows, computing the cosine similarity against the \texttt{audio\_class} text prompt. The window yielding the highest similarity score is extracted.
\item \textbf{Extension via Looping (Duration $>$ 10s):} For extended environmental sounds, the output is seamlessly looped and crossfaded until the target duration is met.
\end{itemize}

\subsection{Auditing and Human-in-the-Loop Refinement}
To maximize acoustic coverage, the cue manifest undergoes a secondary LLM auditing pass. This ``Gap-Filler'' agent cross-references the generated tracks against the original text to synthesize missed auditory opportunities (e.g., adding a sharp gasp SFX). The completed manifest is then exposed via an interactive interface for personalized, fine-grained adjustments prior to superimposition.

\section{The Superimposition and Alignment Engine}
\label{sec:superposition}
Once the individual acoustic stems are generated, the framework must composite them into a unified audio track. The primary challenge is the \textit{Alignment Problem} -- predicting the precise start time ($t_{start}$) and duration ($d_{dur}$) of each audio cue relative to the narrative flow.

Unlike isolated sound generation, cinematic orchestration requires contextual foresight. To solve this, our Superimposition Engine utilizes a hybrid alignment approach, grounded by high-resolution text-to-speech transcription and trained on real-world cinematic data.

\subsection{Temporal Grounding: Whisper JSON}
To establish a deterministic master clock, we transcribe the generated TTS narration using a Whisper-based Automatic Speech Recognition (ASR)~\cite{whisper} model. This generates a granular JSON manifest containing word-level timestamps. For our running horror example, Whisper extracts the exact millisecond timestamps for critical vocal triggers such as ``gasping'' and ``lunged,'' replacing the initial mathematical reading-speed estimations.

\subsection{Cinematic Ground Truth: The YouTube Trailer Dataset}
To train our predictive models on professional pacing, we curated a novel dataset of cinematic audio timings, extracting audio and video from approximately 100 high-quality movie trailers\footnote{Dataset available under the following link:\\
\url{https://docs.google.com/spreadsheets/d/1QyqbvlgpzJ0clwwLy8J5hq-WthJ6IBAN}}, subdivided into 1000 distinct clips. Using a multimodal LLM (Gemini), we performed detailed scene analysis to categorize sounds (SFX, Ambience, Music), logging descriptions, timings, and relative mix volumes. This dataset perfectly encapsulates the editorial rhythm of professional directors—specifically demonstrating how background music consistently anticipates on-screen action to build tension.

\subsection{Digital Sound Predictor (DSPred): Fine-Tuning for Cinematic Orchestration}
We fine-tune a LLaMA-based language model, establishing our Digital Sound Predictor (DSPred). Because the ground-truth clips possess a fixed duration ($D_{clip}$), while our generated TTS narrations exhibit dynamic lengths ($D_{nar}$), we map the dataset's audio cues onto the generated narrator's timeline:
$$t'_{start}=\left(\frac{t_{start}}{D_{clip}}\right)\times D_{nar} \qquad d'_{dur}=\left(\frac{d_{dur}}{D_{clip}}\right)\times D_{nar}$$
We formulate the parameter prediction as a conditional generative task, optimized via Cross-Entropy loss. To maintain computational efficiency, we apply Low-Rank Adaptation (LoRA)~\cite{lora} approach during backpropagation.

\subsection{Hybrid Alignment Strategies}
To accommodate the varied nature of cinematic sounds, the Superimposition Engine resolves the alignment problem using a dynamically weighted combination of methods:
\begin{enumerate}
\item \textbf{LLM-Heuristic Alignment:} Prompts the base LLM with strict alignment rules. Best suited for discrete, synchronous Sound Effects (SFX).
\item \textbf{DSPred Alignment:} Leverages trailer-data training to intuitively build context. Best suited for Musical Scores and Ambience.
\item \textbf{Semantic Adaptive Mixing:} Introduces a semantic weighting parameter, $\alpha$, derived from a Sentence Transformer. The final alignment is calculated as a linear interpolation:
$$t'_{final} = \alpha \cdot t_{LLM} + (1 - \alpha) \cdot t_{DSPred}$$
If the acoustic prompt is semantically closer to sudden actions (SFX), $\alpha$ approaches 1. If it aligns with emotional descriptors (Music), $\alpha$ approaches 0.
\end{enumerate}

\begin{example} {\bf [Alignment Resolution]} \label{ex:alignment}
{\em
Applying the above hybrid logic to our horror-story prompt introduced in Example~\ref{ex:semantic} reveals a critical advantage of DSPred. A naive LLM strictly calculates timestamps based on text triggers, rigidly placing both the tense rising strings and the jump-scare impact SFX at the exact Whisper timestamp for ``lunged''. However, our Semantic Adaptive Mixing differentiates the stems. Because the impact is an SFX, it correctly remains anchored to the word ``lunged''. Conversely, the rising strings are classified as Music, DSPred, having learned from trailer data that music anticipates action, and taking average of both predictions actively shifts the start time of the rising strings several seconds \textit{before} the word ``lunged''. This creates a natural, cinematic tension build-up prior to the climax. \qed
}
\end{example}

\subsection{Final Acoustic Rendering and Signal Processing}
Once mathematically aligned, the audio arrays are superimposed. To prevent amplitude clipping, the system executes automated audio ducking -- dynamically applying a compression threshold to reduce the decibel level of the Ambience and Music tracks whenever the Narrator exhibits vocal activity. Finally, crossfade modulations and volume smoothing algorithms are applied to the overlapping boundaries, rendering a cohesive, broadcast-ready cinematic audio file.

\section{Evaluation Metrics and Methods}
\label{sec:eval_metric}
Evaluating superimposed, multi-track audio without direct ground-truth references remains a significant challenge in generative acoustics. Standard metrics, such as Fréchet Audio Distance (FAD)~\cite{fad} or global CLAP scores~\cite{clap}, primarily assess isolated acoustic quality or global semantic fit, failing to evaluate precise temporal orchestration. To address this, we propose a novel Nearest-Neighbor Retrieval Evaluation framework leveraging our curated YouTube cinematic dataset, alongside a comprehensive human-in-the-loop subjective evaluation platform.

\subsection{Retrieval-Based Objective Metrics}
Our objective evaluation pipeline compares a generated narrative soundscape with the closest real-world cinematic equivalent. Given a generated story prompt $s_{gen}$ and a set of generated audio cues $G = \{g_1, g_2, \dots, g_m\}$, we compute the cosine similarity between the text embedding of $s_{gen}$ and all stories in our dataset using a pre-trained CLAP model. If the highest similarity score exceeds a semantic threshold $\tau_{story}$, we retrieve the closest matching ground-truth story $s_{yt}$ and its associated professional audio cues $A = \{a_1, a_2, \dots, a_n\}$.

We then evaluate the alignment between $G$ and $A$ using two proposed metrics:
\begin{enumerate}
\item \textbf{YT Coverage Score (Semantic Recall):} We evaluate whether the generative model successfully synthesized the necessary acoustic elements of the scene. A generated cue $g_i$ and a ground-truth cue $a_j$ form a matched pair if their textual CLAP embeddings exceed a similarity threshold:
$$\cos(E_{CLAP}(g_i), E_{CLAP}(a_j)) > \tau_{cue}$$
Let $P$ represent the set of all successfully matched pairs. The coverage score is calculated as the recall of the ground-truth cues:
$$\text{Coverage Score} = \frac{|P|}{|A|}$$
Crucially, this recall-based formulation ensures that the generative model is not penalized for additional, contextually appropriate soundscapes beyond what existed in the original YouTube clip.

\item \textbf{YT Sync Score (Temporal IoU):} To evaluate the rhythmic and cinematic pacing of the generated music and environmental sounds, we calculate the 1-Dimensional Intersection Over Union (IoU) for all matched pairs in $P$. For a given pair $(g, a)$, let $S$ denote the start time and $E$ denote the end time. The temporal alignment is calculated as:
$$\text{IoU}(g, a) = \frac{\max(0, \min(E_g, E_a) - \max(S_g, S_a))}{\max(E_g, E_a) - \min(S_g, S_a)}$$
The final Sync Score is the summation of the IoU across all matched pairs, normalized by $|P|$. This heavily penalizes models that trigger background music or effects out of sync with the professional directorial pacing found in the dataset.
\end{enumerate}

\subsection{Human-in-the-Loop Subjective Evaluation}
Because algorithmic metrics often fail to capture the subjective emotional resonance and ``feel'' of a cinematic mix, we developed an open-source web platform to collect granular human perceptual scores. Data is aggregated via a serverless pipeline into a relational database, generating a rich dataset of story-to-audio human preference mappings.

The subjective feedback is systematically segregated into two broad domains:
\begin{enumerate}
\item \textbf{Specialist Level (Unit Testing):} This evaluates the performance of the individual generation models (e.g., Tango2, Parler-TTS) on isolated audio stems. The metrics include Prompt Fidelity, Acoustic Naturalness, and Recognition Rate.
\item \textbf{Master Level (Superimposition Testing):} This evaluates the orchestration capabilities of final mixed audio track. The users score the soundscape on Sync Accuracy, Semantic Fit, Narrative Flow, and overall Cinematic Impact.
\end{enumerate}
This dual-tiered evaluation ensures that both the raw acoustic fidelity of the specialist models and the mixture and alignment logic of the Master Agent are rigorously validated against human acoustic perception.





%% file: sections/5_results_and_evaluation.tex
\section{Experimental Results}
\label{sec:result}

\subsection{Experimental Setup and Ablation Configurations}
To evaluate the efficacy of the BackgroundMellow framework, we sampled ~40 diverse story prompts from our curated YouTube Trailer Dataset. We conducted an extensive ablation study to isolate the impact of our architectural components by toggling the following system parameters:

\begin{enumerate}
    \item \textbf{LLM Backbone (B1):} Comparing base LLM reasoning capabilities (e.g., Gemini-3.0 vs. Gemini-2.5).
    \item \textbf{Auditing Agent / Gap-Filler (B2):} Enabling or disabling the secondary LLM pass for filling missing acoustic coverage.
    \item \textbf{Alignment Strategy:} Comparing naive timestamps against our proposed superimposition methods: (B7) Strict LLM-Heuristic, (B6) DSPred Only, (B0, B3) Unweighted Average (trained/untrained DSPred), (B4) Semantic Adaptive Mixing ($\alpha$-weighted), and (B5) omitting the predictor entirely.
    \item \textbf{Movie BGM Retrieval (B8):} Toggling the Cinematic BGM Retriever versus relying purely on synthesized background music.
    \item \textbf{Narrator (B9):} Disabling the narrator voice to evaluate purely background acoustic stems.
\end{enumerate}

\begin{figure*}[t]
    \centering

    \begin{subfigure}[t]{0.49\textwidth}
        \centering
        \includegraphics[width=\linewidth, height=0.20\textheight]{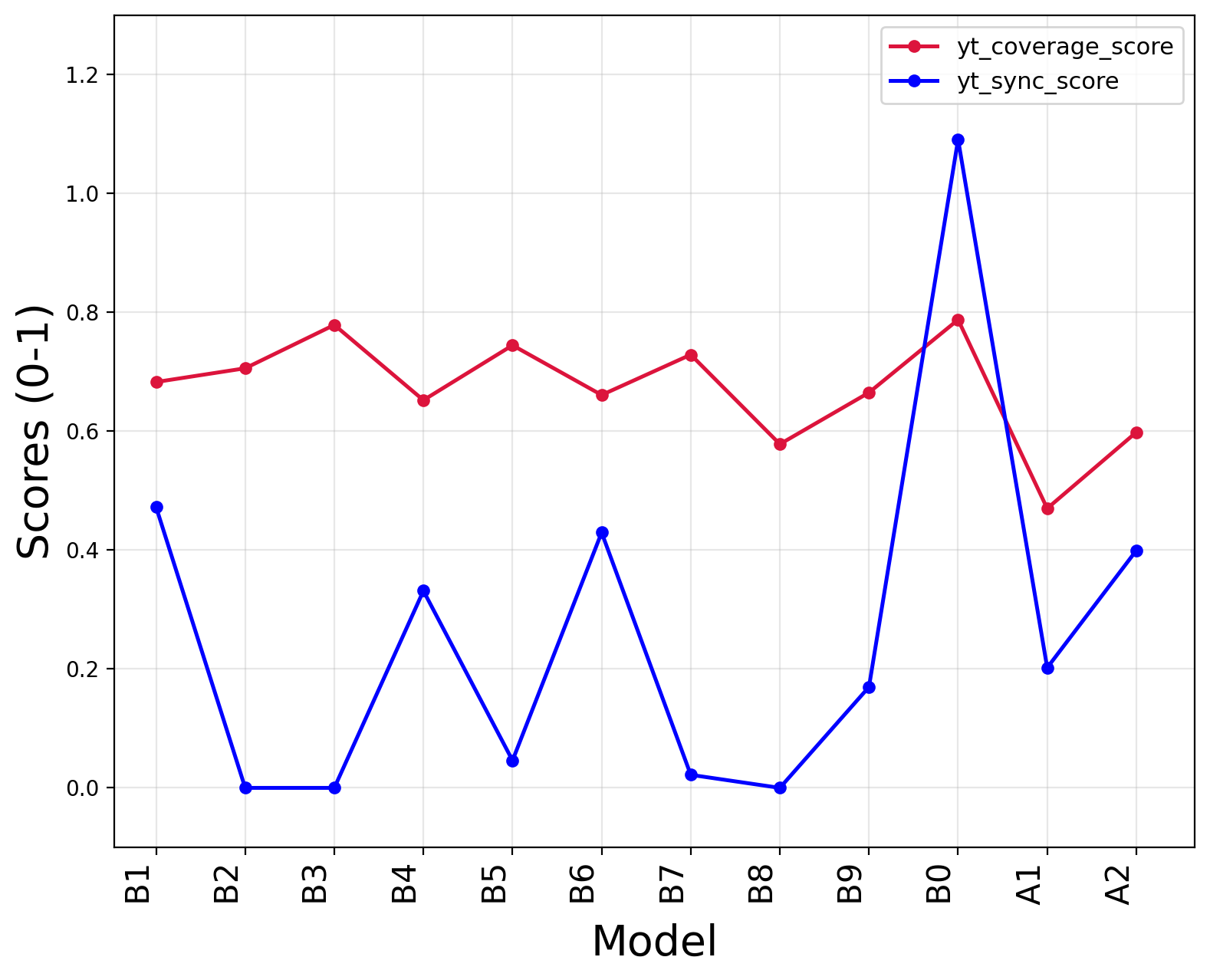}
        \caption{YouTube Metrics}
        \label{fig:yt_scores}
    \end{subfigure}
    \hfill
    \begin{subfigure}[t]{0.49\textwidth}
        \centering
        \includegraphics[width=\linewidth, height=0.20\textheight]{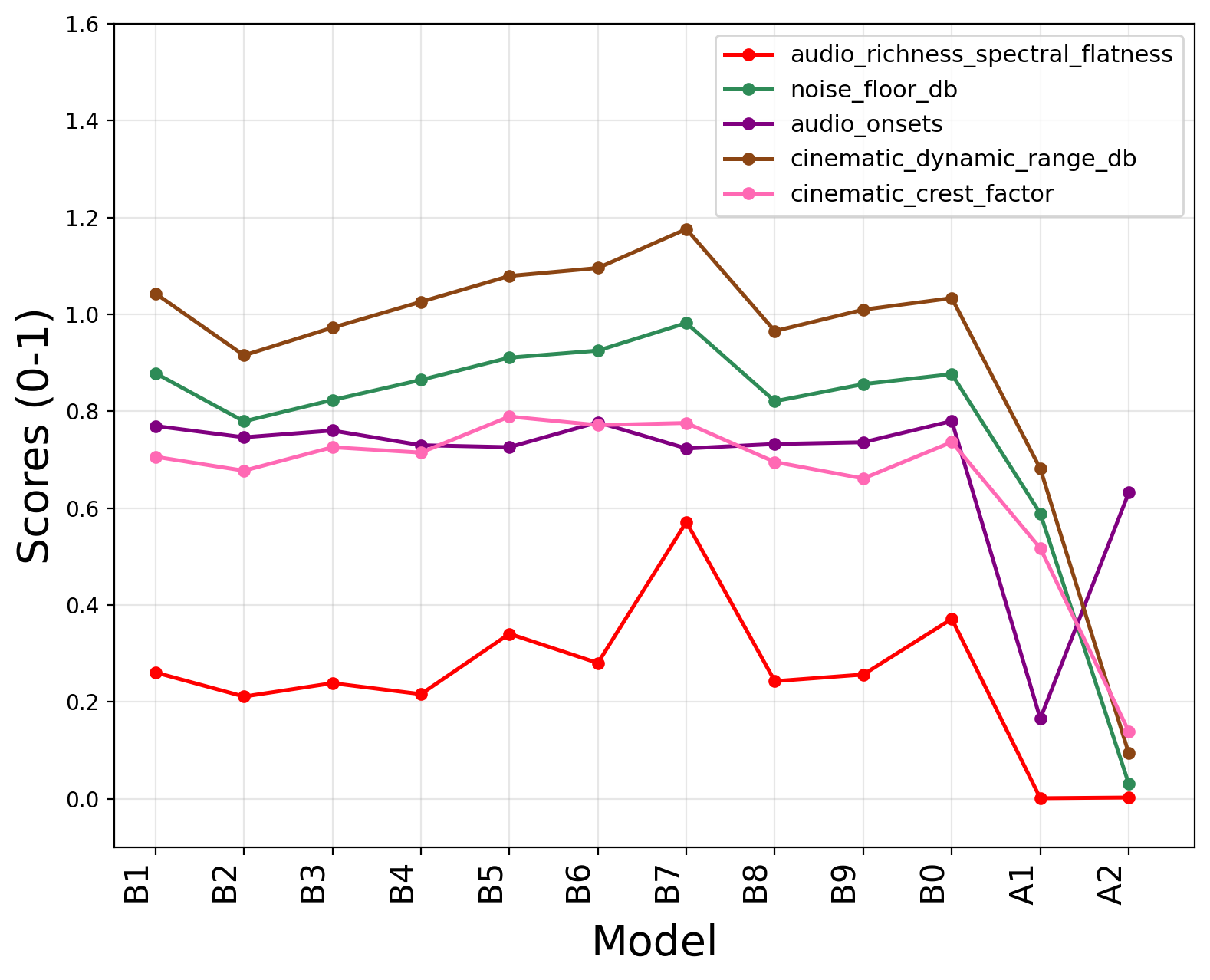}
        \caption{Ablation Results}
        \label{fig:ablations_results}
    \end{subfigure}
    \hfill
    \caption{Comparative Visualization of Scores and Ablation Results across Design Configurations}
    \label{fig:combined}
\end{figure*}

\subsection{Objective Metrics Definition}
To comprehensively assess the generated audio, we define the following acoustic and semantic metrics. For unified graphical visualization in Fig.~\ref{fig:ablations_results}, the absolute values were scaled by the designated uniform factors.

\begin{description}
    \item[Audio Spectral flatness:] an audio evaluation metric that quantifies how ``noise-like'' or ``tone-like'' a sound is by analyzing the distribution of power across its frequency spectrum. The values are scaled by $10$.
    \item[Noise Floor (dB):] Quantifies the baseline level of continuous background noise present in audio track. The values are scaled by $-1/50$.
    \item[Audio Onsets:] Captures the temporal locations of sudden bursts of acoustic energy, indicating rhythmic and transient events (e.g., SFX). The values are scaled by $1/100$.
    \item[Cinematic Dynamic Range (dB):] Measures the decibel difference between the loudest peaks and the quietest segments, reflecting the track's volumetric depth. The values are scaled by $1/100$.
    \item[Cinematic Crest Factor:] Computes the ratio of peak signal amplitude to the root mean square (RMS) value, highlighting the ``punchiness'' of transient peaks. The values are scaled by $1/15$.
\end{description}

\subsection{Performance Analysis and Results}
The quantitative results of our ablation study are presented in Table \ref{tab:configuration_ablations} and visualized in Fig.~\ref{fig:yt_scores} and Fig.~\ref{fig:ablations_results}. 

\begin{table*}[t]

\centering
\caption{Quantitative ablation results demonstrating the impact of design choices. Configuration IDs (A1-A2, B0-B9) map to the components tested in our framework. Best values for each metric are highlighted in \textbf{bold}. Up arrows ($\uparrow$) indicate higher is better, down arrows ($\downarrow$) indicate lower (quieter) is better.}
\label{tab:configuration_ablations}
\resizebox{\textwidth}{!}{%
\begin{tabular}{@{}l ccccccc@{}}
\toprule
\textbf{Configuration} & \textbf{Flatness} $\uparrow$ & \textbf{Noise Floor (dB)} $\downarrow$ & \textbf{Audio Onsets} $\uparrow$ & \textbf{Dyn. Range (dB)} $\uparrow$ & \textbf{Crest Factor} $\uparrow$ & \textbf{YT Cov} $\uparrow$ & \textbf{YT Sync} $\uparrow$ \\ 
\midrule

\multicolumn{8}{@{}l}{\textit{Monolithic Baselines}} \\ 
\midrule
\textbf{A1}: AudioLDM2 (Zero-Shot)       & 0.000 & -29.459 & 16.6 & 68.153 & 7.761 & 14.11 & 1.01e-03 \\ 
\textbf{A2}: Tango2 (Zero-Shot)          & 0.000 & -1.548  & 63.2 & 9.406  & 2.090 & 17.95 & 1.99e-03 \\

\midrule
\multicolumn{8}{@{}l}{\textit{\bgm~Framework (Ours)}} \\ 
\midrule
\textbf{B0: Baseline (Avg DSP,LLM Align)} 
& 0.037 & -43.839 & \textbf{78.0} 
& 103.368 & 11.048 
& \textbf{23.635} & \textbf{5.46e-03} \\

\textbf{B1}: Gemini 2.5 Backbone        
& 0.026 & -43.939 & 77.0 
& 104.329 & 10.592 
& 20.490 & 2.36e-03 \\

\textbf{B2}: + Gap-Filler Coverage     
& 0.021 & -38.963 & 74.6  
& 91.563  & 10.161 
& 21.183 & $\approx 0$ \\

\textbf{B3}: UnTrained DSP and LLM Avg Align
& 0.024 & -41.169 & 76.0  
& 97.274  & 10.888 
& 23.372 & $\approx 0$ \\

\textbf{B4}: + DL Adaptive Align     
& 0.022 & -43.239 & 73.0  
& 102.598 & 10.721 
& 19.572 & 1.66e-03 \\

\textbf{B5}: - Alignment Predictor       
& 0.034 & -45.540 & 72.6 
& 107.931 & \textbf{11.838} 
& 22.344 & 2.30e-04 \\

\textbf{B6}: DSPred Align Only 
& 0.028 & -46.267 & 77.7  
& 109.584 & 11.573 
& 19.833 & 2.15e-03 \\

\textbf{B7}: LLM Align Only 
& \textbf{0.057} & \textbf{-49.118} & 72.3  
& \textbf{117.623} & 11.639 
& 21.869 & 1.10e-04 \\

\textbf{B8}: + Movie BGMs 
& 0.024 & -41.040 & 73.3  
& 96.569  & 10.430 
& 17.353 & $\approx 0$ \\

\textbf{B9}: - Narrator 
& 0.026 & -42.805 & 73.6  
& 100.979 & 9.919 
& 19.955 & 8.49e-04 \\

\bottomrule
\end{tabular}%
}
\end{table*}

\begin{description}[leftmargin=0pt]
\item[{\bf Acoustic Quality vs. Monolithic Compression:}]
Our experiments reveal the fundamental limitations of monolithic Text-to-Audio models. Zero-shot Tango2 exhibits severe dynamic compression (a ``brick-wall'' effect) with a dynamic range of only $9.41$~dB and an unnatural $-1.55$~dB noise floor. Conversely, \bgm~maintains a pristine $103.37$~dB dynamic range and an $11.05$ Crest Factor. This mathematically validates that our explicit DSP ducking preserves acoustic fidelity across overlapping stems, avoiding the flat, noisy mixtures of end-to-end models.
\smallskip

\item[{\bf Narrative Density and Cinematic Glue:}]
AudioLDM2 generates sparse soundscapes with only $16.6$ audio onsets, whereas our framework yields $78.0$ onsets, accurately reflecting discrete narrative action triggers (SFX). Furthermore, integrating the Cinematic BGM Retriever naturally compresses the dynamic range to $96.57$~dB, demonstrating that retrieved professional scores effectively act as ``cinematic glue,'' filling digital silence without overpowering the narrator.
\smallskip

\item[{\bf YouTube Coverage and Temporal Synchronization:}]
To fairly evaluate the monolithic baselines (Tango2~\cite{tango2} and AudioLDM2~\cite{audioldm2}) using our proposed retrieval-based metrics, we performed an audio-to-text reverse-engineering step. Because end-to-end models output raw waveforms without cue metadata, we utilized a multimodal LLM (Gemini) to perform detailed scene analysis on their generated audio. The LLM extracted the constituent sounds, start times, and durations to create a comparative cue manifest. We then computed the \texttt{YT Coverage} and \texttt{YT Sync} scores against our ground truth, mirroring our framework's evaluation methodology.

The experimental data strongly validates our decoupled orchestration approach. Monolithic models struggled to concurrently synthesize all necessary narrative elements, resulting in lower \texttt{YT Coverage} scores ($17.95$ for Tango2 and $14.11$ for AudioLDM2). In contrast, the BackgroundMellow baseline achieved a robust coverage score of $23.63$. This proves that the Master Agent effectively parses and delegates complex textual prompts into dense, multi-layered acoustic scenes that more accurately reflect real-world cinematic soundscapes.
\end{description}

%% file: sections/6_future_and_conclusion.tex
\section{Conclusion}
\label{sec:conclusion}
The \bgm~framework demonstrates that high-fidelity narrative sound design necessitates explicit architectural orchestration rather than monolithic generation. By decoupling semantic planning from audio synthesis by a Master-Agent paradigm with specialized generators, augmented BGM retrieval, and DSPred-based temporal alignment, the system achieves precise and controllable audio superimposition. Overall, BackgroundMellow~establishes a foundational framework for transforming textual narratives into coherent, temporally aligned cinematic soundscapes, moving toward fully autonomous audio direction. In future, we plan to involve a Generative Adversarial-style refinement loop, where a perceptual discriminator iteratively guides the Master Agent to refine mix parameters based on human satisfaction scores.

%% file: sections/7_references.tex

%% file: main.bbl
\begin{thebibliography}{99}

\bibitem{audiostory}
Yuxin Guo, Teng Wang, Yuying Ge, Shijie Ma, Yixiao Ge, Wei Zou, and Ying Shan. 
\newblock ``AudioStory: Generating Long-Form Narrative Audio with Large Language Models.'' 
\newblock \emph{https://arxiv.org/abs/2508.20088}, 2025.

\bibitem{audiobook_agent}
Arul Selvamani Shaja, PhD1 and Nia D’Souza Ganapathy2
\newblock ``A Multi-Agent AI Framework for Immersive Audiobook Production through Spatial Audio and Neural Narration.'' 
\newblock \emph{https://arxiv.org/pdf/arXiv:2505.04885}, 2025.

\bibitem{tango}
Deepanway Ghosal, Navonil Majumder, Ambuj Mehrish, and Soujanya Poria. 
\newblock ``Text-to-Audio Generation using Instruction-Tuned LLM and Latent Diffusion Model.'' 
\newblock \emph{https://arxiv.org/pdf/2304.13731}, 2023.

\bibitem{tango2}
Navonil Majumder, Chia-Yu Hung, Deepanway Ghosal, Wei-Ning Hsu, Rada Mihalcea, and Soujanya Poria. 
\newblock ``Tango 2: Aligning Diffusion-based Text-to-Audio Generations through Direct Preference Optimization.'' 
\newblock \emph{https://arxiv.org/pdf/2404.09956}, 2024.

\bibitem{audioldm2}
Haohe Liu, Yi Yuan, Xubo Liu, Xinhao Mei, Qiuqiang Kong, Qiao Tian, Yuping Wang, Wenwu Wang, Yuxuan Wang, and Mark D. Plumbley. 
\newblock ``AudioLDM 2: Learning holistic audio generation with self-supervised pretraining.'' 
\newblock \emph{https://arxiv.org/pdf/2308.05734}, 2024.

\bibitem{ddpm}
Jonathan Ho, Ajay Jain, and Pieter Abbeel. 
\newblock ``Denoising diffusion probabilistic models.'' 
\newblock \emph{https://arxiv.org/pdf/2006.11239}, 2020.

\bibitem{musicgen}
Jade Copet, Felix Kreuk, Itai Gat, Tal Remez, David Rozenberszki, Gabriel Synnaeve, Yossi Adi, and Alexandre Défossez. 
\newblock ``Simple and Controllable Music Generation.'' 
\newblock \emph{https://arxiv.org/pdf/2306.05284}, 2023.

\bibitem{musiclm}
Andrea Agostinelli, Timo I. Denk, Zalán Borsos, Jesse Engel, Mauro Verzetti, Antoine Caillon, Qingqing Huang, Aren Jansen, Adam Roberts, Marco Tagliasacchi, et al. 
\newblock ``MusicLM: Generating Music From Text.'' 
\newblock \emph{https://arxiv.org/pdf/2301.11325}, 2023.

\bibitem{lora}
Edward J. Hu, Yelong Shen, Phillip Wallis, Zeyuan Allen-Zhu, Yuanzhi Li, Shean Wang, Lu Wang, and Weizhu Chen. 
\newblock ``LoRA: Low-Rank Adaptation of Large Language Models.'' 
\newblock \emph{https://arxiv.org/pdf/2106.09685}, 2022.

\bibitem{m2m_gen}
Megha Sharma, Muhammad Taimoor Haseeb,Gus Xia, Yoshimasa Tsuruoka
\newblock ``M2M-Gen: A Multimodal Framework for Automated Background Music Generation in Japanese Manga Using Large Language Models.'' 
\newblock \emph{https://arxiv.org/pdf/2410.09928}, 2024.

\bibitem{metabgm}
\newblock ``MetaBGM: Dynamic Soundtrack Transformation For Continuous Multi-Scene Experiences With Ambient Awareness And Personalization.'' 
\newblock \emph{https://arxiv.org/pdf/arXiv:2409.03844}, 2024.

\bibitem{align_evaluation}
Yichen Huang, Zachary Novack, Koichi Saito, Jiatong Shi, Shinji Watanabe, Yuki Mitsufuji, John Thickstun, and Chris Donahue. 
\newblock ``Aligning Text-to-Music Evaluation with Human Preferences.'' 
\newblock \emph{https://arxiv.org/pdf/arXiv:2503.16669}, 2025.

\bibitem{bach2bach}
Nikhil Kotecha
\newblock ``Bach2Bach: Generating Music Using A Deep Reinforcement Learning Approach.'' 
\newblock \emph{https://arxiv.org/pdf/1812.01060}.

\bibitem{whisper}
Alec Radford, Jong Wook Kim, Tao Xu, Greg Brockman, Christine McLeavey, and Ilya Sutskever. 
\newblock ``Robust speech recognition via large-scale weak supervision.'' 
\newblock \emph{https://arxiv.org/pdf/2212.04356}, PMLR, 2023.

\bibitem{clap}
Yusong Wu, Ke Chen, Tianyu Zhang, Yuchen Hui, Taylor Berg-Kirkpatrick, and Shlomo Dubnov. 
\newblock ``Large-scale contrastive language-audio pretraining with feature fusion and keyword-to-caption augmentation.'' 
\newblock \emph{https://arxiv.org/pdf/2211.06687}, 2023.

\bibitem{fad}
Kevin Kilgour, Mauricio Zuluaga, Dominik Roblek, and Matthew Sharifi. 
\newblock ``Fréchet Audio Distance: A Reference-Free Metric for Evaluating Music Enhancement Algorithms.'' 
\newblock \emph{https://arxiv.org/pdf/1812.08466}, 2019.



\bibitem{parler-tts}
Dan Lyth, Simon King, Stability AI 
\newblock ``Natural language guidance of high-fidelity text-to-speech with synthetic
annotations'' 
\newblock \emph{https://arxiv.org/pdf/2402.01912}, 2017.

\end{thebibliography}
